# Explanation Trees for Causal Bayesian Networks

**Ulf H. Nielsen**[1] uln@zurich.ibm.com
**Jean-Philippe Pellet**[1,2] jep@zurich.ibm.com
**André Elisseeff**[1] ael@zurich.ibm.com

[1] Business Optimization Group
IBM Research GmbH
8803 Rüschlikon, Switzerland

[2] Machine Learning Group
Swiss Federal Institute of Technology Zurich
8092 Zurich, Switzerland

## Abstract

Bayesian networks can be used to extract explanations about the observed state of a subset of variables. In this paper, we explicate the desiderata of an explanation and confront them with the concept of explanation proposed by existing methods. The necessity of taking into account causal approaches when a causal graph is available is discussed. We then introduce causal explanation trees, based on the construction of explanation trees using the measure of causal information flow (Ay and Polani, 2006). This approach is compared to several other methods on known networks.

## 1 INTRODUCTION

A Bayesian network (BN, Pearl, 1988) is an algebraic tool to compactly represent the joint probability distribution of a set of variables $\mathbf{V}$ by exploiting conditional independence amongst variables. It represents all variables in a directed acyclic graph (DAG), where the absence of arcs between nodes denotes (conditional) independence. In addition to graphically representing the structure of the dependencies between the variables, BNs allow inference tasks to be solved more efficiently. In this paper, we discuss the extraction of explanations in *causal BNs* (Pearl, 2000; Spirtes et al., 2001)—BNs where the arcs depict direct cause–effect relationships between variables.

Generally, explanations in BNs can be classified in three categories (Lacave and Diez, 2002) depending on the focus of the explanation:

- Explanation of *evidence*. Given a subset of observed (instantiated) variables $\mathbf{O} \subsetneq \mathbf{V}$, what is the state of (some of) the other variables $\mathbf{V} \setminus \mathbf{O}$ that best explains $\mathbf{O} = \mathbf{o}$?
- Explanation of the *reasoning process*. When we have received some evidence and belief states are updated by probabilistic inference, how was the reasoning process by which we arrive at this state?
- Explanation of the *model*, which provides insight into the static components of a network such as (conditional) independence relationships, causal mechanisms, etc.

We shall focus our attention on the explanation of evidence: we wish to explain why variables in $\mathbf{O}$ took on specific observed values using assignments in $\mathbf{V} \setminus \mathbf{O}$. To this purpose, we discuss in section 2 the requirements of such an explanation. In section 3, we list the standard approaches to evidence explanation as well as some recent methods to make explanations more concise, and explain some of their drawbacks. We then present causal information trees in section 4, and detail experiments and comparisons in section 5.

### NOTATION

Boldface capitals denote sets of random variables or nodes in a graph, depending on the context. $\mathbf{V}$ is the set of all variables in the analysis. Italicized capitals like $X$, $X_i$, $Y$ are random variables or nodes and elements of $\mathbf{V}$; calligraphic capitals such as $\mathcal{X}, \mathcal{Y}$ are their respective domains. Vectors are denoted boldface lowercase, as $\mathbf{e}$ or $\mathbf{p}$; scalars in italics. Unless otherwise stated, the scalars $x, y$ are assumed to be a value of their respective uppercase random variable. The probability distribution of a random variable $X$ is denoted by $p(X)$, and we write $p(X = x)$ or $p(x)$ the probability of $x$. We only work with discrete variables.

## 2 AN IDEALIZED EXPLANATION

Even though many variables $\mathbf{O}$ may be observed, the explanation can be focused only on a specific subset $\mathbf{E} \subseteq \mathbf{O}$. The state $\mathbf{E} = \mathbf{e}$ is then called *explanandum*.

The set of explanatory variables $\mathbf{H} \subsetneq \mathbf{V}$ can include both observed and unobserved variables, and an explanation is an assignment $\mathbf{H} = \mathbf{h}$ (compatible with $\mathbf{O} = \mathbf{o}$ for variables both in $\mathbf{H}$ and in $\mathbf{O}$).

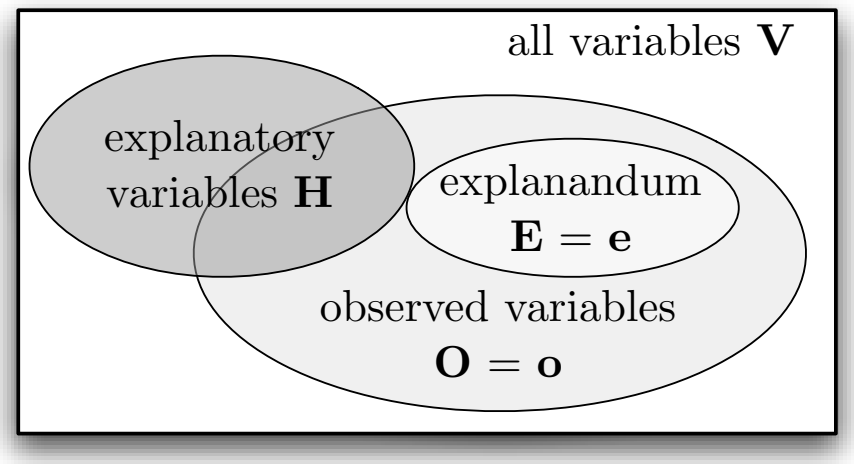

We insist on the distinction between the explanandum $\mathbf{e}$ and the observations $\mathbf{o}$ (Chajewska and Halpern, 1997). Observations are all our knowledge about the current state of a system, and this might not coincide exactly with what we want explained. Consider for example the case where we wish to know why the grass is wet while we know it has been raining. We do not seek an explanation for why it has rained, only for why the grass is wet. A perfectly valid explanation is that the grass is wet because it is raining if no other factors can sufficiently explain the facts.

An algorithm respecting this should then determine, for each variable in $\mathbf{O}$, whether its observed state is relevant to explain $\mathbf{e}$, and for each unobserved variable in $\mathbf{V} \setminus \mathbf{O}$, whether knowing its state adds "explanatory power" to the proposed explanation. This excludes methods which marginalize out $\mathbf{O}$, preventing these variables from being part of an explanation.

To explain *why* a given system is observed in a given state, we must intuitively convey some information about the causal mechanisms that lead to the observation made. If we observe that it is raining and some explanation tells us that "it rains because the grass is wet," we do not find it a good explanation as it contradicts our understanding of how the system works; that the grass being wet cannot make it rain. Suppose we have an explanation $\mathbf{H} = \mathbf{h}$ for $\mathbf{E} = \mathbf{e}$: an intuitive interpretation of this result is that manually setting $\mathbf{H} = \mathbf{h}$ will be a favourable configuration to observe $\mathbf{E} = \mathbf{e}$. As Halpern and Pearl (2005) discuss, explanations need to be causal to be consistent with users' knowledge of the mechanisms of the system. It is therefore important that the explanations are given in a data-generating direction, such that users can infer interventional rules from the given explanations (for instance, "if I can make it rain somehow, then I know that the grass will be wet," as opposed to an impossible "let me make the grass wet so as to make it rain").

Causal methods are subject to the availability of causal information. In this paper, we extract the causal information from causal BNs, but in general, the approach is adaptable to any causal model that can predict the effect of interventions on certain variables.

In addition to assuming that the relationships between the variables $\mathbf{V}$ can be represented by a fully oriented causal BN, we assume that the corresponding joint probability distribution is faithful and causally sufficient (Pearl, 2000; Spirtes et al., 2001). *Faithfulness* of the distribution ensures that there is a unique graph whose arcs depict all (conditional) dependencies of the distribution, and only those. *Causal sufficiency* forbids hidden common causes for variables in $\mathbf{V}$, such that we can build a DAG whose arcs represent direct causation. Although most expert-designed BN are naturally oriented causally, the output of structure causal learning algorithms are often partially directed graphs and may need additional expert knowledge to be fully oriented.

To summarize, we wish our explanations to give us causal information by detailing the mechanisms that lead to the explanandum, using all the available information we have about the state of the network.

# 3 EXISTING METHODS

This section reviews and discusses some of the major techniques to find explanations.

## 3.1 MOST PROBABLE EXPLANATION & VARIANTS

A common noncausal measure of explanatory power is the conditional probability of the explanatory variables $\mathbf{H}$ given the explanandum $\mathbf{e}$. The *most probable explanation* (MPE) approach (Pearl, 1988) then considers $\mathbf{h}^* = \arg\max_{\mathbf{h}} p(\mathbf{h} \,|\, \mathbf{e})$ as the best explanation (or, alternatively, looks for the $k$ best explanations by maximizing this probability). The explanandum $\mathbf{e}$ is in the case of MPE equal to the full set of observations $\mathbf{O} = \mathbf{o}$, and the set $\mathbf{H}$ is $\mathbf{V} \setminus \mathbf{E}$. This list can be long and uninformative because of lack of conciseness; moreover, it is hard to distinguish between long explanations, whose respective probabilities are low anyway and close to one another.

In the *partial abduction* approach (Shimony, 1991), the set of explanatory variables is a strict subset $\mathbf{H} \subsetneq \mathbf{V} \setminus \mathbf{E}$. The set of variables $\mathbf{X} = \mathbf{V} \setminus \mathbf{H} \setminus \mathbf{E}$ excluded from the explanation is then marginalized out before the maximum is computed: we look for $\arg\max_{\mathbf{h}} \sum_{\mathbf{x}} p(\mathbf{h}, \mathbf{x} \,|\, \mathbf{e})$. This is the *maximum a posteriori* (MAP) model approach. The excluded variables $\mathbf{X}$ are selected either by a user, or via automated analysis of the network. Automatically selecting the relevant explanatory variable is a nontrivial issue (Shimony, 1991).

Partial abduction is computationally more expensive than standard MPE, because it cannot be readily solved by message passing algorithms, but approximations exist (e.g., Park, 2002). On the other hand, it globally leads to more concise explanations than MPE.

Further efforts to make explanations more concise include de Campos et al. (2001), where the $k$ most probable explanations are found and then simplified based on relevance and probabilistic criteria; and Henrion and Druzdzel (1990), where also partial assignments are allowed but only within a predefined tree that limits the set of possible explanations. An explanation is then a path from the root of the tree to a leaf, denoting variable assignments for each branch taken. This is known as *scenario-based explanation*. The best explanation is the one with the highest posterior probability.

There are several concerns with these approaches—MPE/MAP or scenario-based—maximizing some conditional probability of the explanatory variables (Chajewska and Halpern, 1997). First, they do not distinguish the explanandum and the observations, such that the additional state information that is not meant to be explained is excluded from a possible explanation. Furthermore, there is no distinction between observing an explanatory variable $X$ in a certain state $x$, and forcing it to have the value $x$.[1] Thus, depending on the choice of the explanatory variables, the intuitive interpretation (as described in the previous section) stating that setting $\mathbf{H} = \mathbf{h}^*$ will be a favourable configuration for observing $\mathbf{E} = \mathbf{e}$ does not hold.

MPEs and, to a lesser extent, MAP model explanations, are not robust: little changes in the network will often change the result of the analysis, even though the changes occur in parts of the network largely independent of the explanandum (Chan and Darwiche, 2006). Common to the methods in this subsection is that they order explanations by $p(\mathbf{h}, \mathbf{e})$ (this is equivalent to $p(\mathbf{h} \mid \mathbf{e})$ as $p(\mathbf{e})$ is constant for a given $\mathbf{e}$): this joint probability cannot be considered alone to determine the explanatory power of $\mathbf{h}$ on $\mathbf{e}$. Some of these problems are illustrated by the experiments in section 5.

### 3.2 SE ANALYSIS

In SE analysis, Jensen (2001) additionally considers the sensitivity of an explanation $\mathbf{h}$ with respect to the explanandum. Less sensitive explanations ensure that little changes in the network's parameters will not lead to severely different explanations, so that the explanation is stable with respect to the specification of the network.

[1] The difference between observation and intervention is fundamental to causality and is best described with the example of Simpson's paradox in Pearl (2000), chap. 6.

SE analysis also works by comparing two explanations $\mathbf{h}_i$ and $\mathbf{h}_j$, usually with *Bayes' factor* or the *likelihood ratio* (Jeffreys, 1961): Bayes' factor =

$$\frac{\text{posterior ratio}}{\text{prior ratio}} = \frac{p(\mathbf{h}_i \mid \mathbf{e}) \,/\, p(\mathbf{h}_j \mid \mathbf{e})}{p(\mathbf{h}_i) \,/\, p(\mathbf{h}_j)} = \frac{p(\mathbf{e} \mid \mathbf{h}_i)}{p(\mathbf{e} \mid \mathbf{h}_j)}.$$

The empirical interpretation of Bayes' factor given by Jeffreys (1961) is that if it is less than 1 it is in favor of $\mathbf{h}_j$, if less than 3 it is a slight support for $\mathbf{h}_i$. If it is between 3 and 12, it is a positive support; and higher than 12, it is a strong support for $\mathbf{h}_i$.

In Yuan and Lu (2007), Bayes' factor is used to search for explanations consisting of only a few variables by ranking them by their Bayes' factor computed as the ratio between the probability of the explanation given the explanandum and its opposite:

$$\text{Bayes' factor} = \frac{p(\mathbf{h} \mid \mathbf{e})}{1 - p(\mathbf{h} \mid \mathbf{e})}.$$

An exhaustive search is performed over all subsets of the hypothesis, and the explanations are shown to be more concise in a sample network than MPE, Shimony's (1991) MAP, and the simplifications described by de Campos et al. (2001).

A similar criticism as before can be applied to these methods: additional observations are discarded, and the causal directionality is ignored in the selection of the relevant explanatory variables.

### 3.3 EXPLANATION TREES

The method of Flores (2005) constructs a set of best explanations while at the same time giving a preference for concise explanations, summarizing the results of the analysis in an *explanation tree*. We describe this method in more detail, as the causal information tree method (described in section 4) is based on a similar representation.

**Definition 1** *An* ***explanation tree*** *for an explanandum* $\mathbf{E} = \mathbf{e}$ *is a tree in which every node* $X$ *is an explanatory variable (with* $X \in \mathbf{V} \setminus \mathbf{E}$*), and each branch out of* $X$ *is a specific instantiation* $x \in \mathcal{X}$ *of* $X$*. A path from the root to a leaf is then a series of assignments* $X = x$, $Y = y$, $\cdots$, $Z = z$, *summarized as* $\mathbf{P} = \mathbf{p}$, *which constitutes a full explanation.*

Flores's (2005) algorithm, summarized in Algorithm 1, builds such an explanation tree. Starting with an empty tree, the variable to use as the next node is selected to be the one that, given the explanandum, reduces the uncertainty the most in the rest of the explanatory variables according to some measure. The nodes that are on the path being grown are added to a conditioning set, so that the part of the explanatory space they already account for is taken into account. Two stopping criteria are used to determine when to stop growing the tree: the minimum posterior probability $\beta$ of the current branch, and the minimum amount of uncertainty reduction $\alpha$ that must be achieved by adding a new variable. Among all explanations represented by the final tree, the best one is the one with the largest posterior probability $p(\mathbf{p} \mid \mathbf{e})$.

**Algorithm 1** Flores's (2005) Explanation Tree

1: **function** $T = \text{ExplanationTree}(\mathbf{H}, \mathbf{e}, \mathbf{p};\ \alpha, \beta)$

| | | |
|---|---|---|
| **Input:** | $\mathbf{H}$ : | set of explanatory variables |
| | $\mathbf{E} = \mathbf{e}$ : | explanandum |
| | $\mathbf{P} = \mathbf{p}$ : | path of variable assignments |
| | $\alpha, \beta$ : | stopping criteria |
| **Output:** | $T$ : | an explanation tree |

2: $X^* \leftarrow \arg\max_{X \in \mathbf{H}} \sum_{Y \in \mathbf{H}} \text{Inf}(X; Y \mid \mathbf{e}, \mathbf{p})$
3: **if** $\max_{Y \in \mathbf{H} \setminus X^*} \text{Inf}(X; Y \mid \mathbf{e}, \mathbf{p}) < \alpha$ **or** $p(\mathbf{p} \mid \mathbf{e}) < \beta$ **then**
4:     **return** $\emptyset$
5: **end if**
6: $T \leftarrow$ new tree with root $X^*$
7: **for each** $x \in \text{domain}(X^*)$ **do**
8:     $T' \leftarrow \text{ExplanationTree}(\mathbf{H} \setminus X^*, \mathbf{e}, \mathbf{p} \cup \{x\})$
9:     add a branch $x$ to $T$ with subtree $T'$ and
10:     assign it the label $p(\mathbf{p}, x \mid \mathbf{e})$
11: **end for**
12: **return** $T$

The algorithm is parametrized with the measure of uncertainty reduction $\text{Inf}(X; Y \mid \mathbf{e}, \mathbf{p})$.[2] For our implementation, we used the *conditional mutual information.* The mutual information $I(X; Y \mid \mathbf{z})$ is a symmetrical measure of how much reduction in uncertainty about $Y$ we get by knowing $X$ in the context $\mathbf{Z} = \mathbf{z}$, and is defined as: $I(X; Y \mid \mathbf{z}) =$

$$\sum_{x \in \mathcal{X}} p(x \mid \mathbf{z}) \sum_{y \in \mathcal{Y}} p(y \mid x, \mathbf{z}) \log \frac{p(y \mid x, \mathbf{z})}{p(y \mid \mathbf{z})}. \qquad (1)$$

If $X$ and $Y$ are independent given $\mathbf{Z} = \mathbf{z}$, we have $I(X; Y \mid \mathbf{z}) = 0$. If $X$ fully determines $Y$, then knowing one is enough to know the other and full information is shared.

Explanation trees are interesting in that they can present many mutually exclusive explanations in a compact form. Flores (2005) also argues that explanations as constructed by Algorithm 1 are reasonable and more sensible than ($k$-)MPE in the sense that on simple networks, the returned explanations are those that we expect. Four elements, however, are subject to discussion.

First, on line 2 of Algorithm 1, variables are added to the tree in order of how much information they provide about the remaining variables in the set of explanatory variables. But this does not measure the information of the added variables shared with the explanandum. Moreover, the explanandum actually grows as the tree is constructed, since there is no difference between the constructed path $\mathbf{p}$ and $\mathbf{e}$ at line 2. Thus, this maximization cannot be interpreted as selecting variables reducing the uncertainty in the explanandum.

Second, the algorithm makes no distinction between explanandum and observations. To try to fix this, we could either additionally condition on observations $\mathbf{O} = \mathbf{o}$, or marginalize out $\mathbf{O}$ altogether. The former case is no different from adding $\mathbf{o}$ to the explanandum $\mathbf{e}$, and the latter case excludes all $X \in \mathbf{O}$ from explanations, such that both cases are unsatisfactory.

Third, the criterion to choose the best explanation is the probability of the explanation path given the evidence $p(\mathbf{p} \mid \mathbf{e})$, and not how likely the system is to have produced the evidence we are trying to explain with a configuration $\mathbf{p}$, $p(\mathbf{e} \mid \mathbf{p})$. Both measures are linked, but since several explanations can cover almost an equal share of the explanation space and often only one will be included in the explanation tree, the criterion $p(\mathbf{p} \mid \mathbf{e})$ will miss explanations which could have explained the evidence well, but do not cover as large a fraction of the explanation space.

Fourth, causal considerations are ignored: there is no distinction between ancestors and descendants of a variables, such that we can get explanations of the type "it rains because the grass is wet."

From an end-user perspective though, trees are a good solution for representing several competing explanations compactly and readably. We introduce in section 4 a modified approach, which can address the issues discussed here.

## 4 CAUSAL EXPLANATION TREES

Like the previous method, causal explanation trees take advantage of a tree representation. The tree is grown so as to ensure that explanations in any path are causal: variables can be selected as explanatory only if they causally influence the explanandum.

Before defining the causal criterion used in this approach, we need to define the concept of *postintervention distribution* (Pearl, 2000, p. 72). A standard conditional probability of the form $p(\mathbf{e} \mid x)$ gives the probability (or probability density) of $\mathbf{e}$ when $X = x$ is observed. It does not represent, however, the probability of $\mathbf{e}$ if we manually force variable $X$ to have value $x$. Causally, we are interested in the *intervention* on $X$, which we denote by $do(X = x)$, rather the observation of $x$. In causal BNs, the tool used to evaluate the effect of these conditionings is Pearl's (1995) *do*-calculus, which uses the structure of the causal graph to evaluate the postintervention distribution.

[2] See Flores (2005) for additional cases where max at line 3 is replaced by min or avg, and Inf is the Gini index.

**Definition 2** *Given a causal Bayesian network $\mathcal{B}$ in the sense of Pearl (2000, p. 23) over variables $\mathbf{V} = \{X_1, \cdots, X_d\}$, the* ***postintervention distribution*** $p(\mathbf{v} \mid do(X_i = x_i'))$*, also denoted* $p(\mathbf{v} \mid \hat{x}_i')$*, after an intervention* $do(X_i = x_i')$ *can be expressed as:*

$$p(x_1, \cdots, x_d \mid \hat{x}_i') = \begin{cases} \prod_{j \neq i} p(x_j \mid \mathbf{pa}_j) & \text{if } x_i = x_i', \\ 0 & \text{if } x_i \neq x_i', \end{cases} \quad (2)$$

*where* $\mathbf{pa}_j$ *is the values of the graphical parents (i.e., direct causes) of the node* $X_j$ *in* $\mathcal{B}$.

The *truncated factorization* of (2) states that the probability distribution is computed as if the manipulated variable $X_i$ had no incoming causal influence (i.e., no direct causes), and as if $p(X_i = x_i')$ had probability one. This makes sense, as forcing $X_i$ to have a certain value effectively ignores its direct causes and "natural" distribution.

With this concept, we can now define the *causal information flow* (Ay and Polani, 2006), which will be our measure of causal contribution of explanatory variables towards our explanandum.

**Definition 3** *The* ***causal information flow*** *from $X$ to $Y$ given the interventions $do(\mathbf{Z} = \mathbf{z})$, written $I(X \to Y \mid \hat{\mathbf{z}})$, is:* $I(X \to Y \mid \hat{\mathbf{z}}) =$

$$\sum_{x \in \mathcal{X}} p(x \mid \hat{\mathbf{z}}) \sum_{y \in \mathcal{Y}} p(y \mid \hat{x}, \hat{\mathbf{z}}) \log \frac{p(y \mid \hat{x}, \hat{\mathbf{z}})}{p^*(y \mid \hat{\mathbf{z}})}, \quad (3)$$

*where* $p^*(y \mid \hat{\mathbf{z}}) = \sum_{x' \in \mathcal{X}} p(x' \mid \hat{\mathbf{z}}) p(y \mid \hat{x}', \hat{\mathbf{z}})$.

The expression $I(X \to Y \mid \hat{\mathbf{z}})$ measures the amount of information flowing from $X$ to $Y$ if we intervene on $\mathbf{Z}$, setting it to $\mathbf{z}$ (i.e., if we block the causal flow on all paths going through $\mathbf{Z}$). Note that (3) is, in essence, similar to (1). For faithful probability distributions, $I(X \to Y \mid \hat{\mathbf{z}}) = 0$ if and only if all directed paths (if any) from $X$ to $Y$ go through $\mathbf{Z}$ in the corresponding causal graph. For binary variables, if $Y$ is a deterministic function of $X$ regardless of $\mathbf{Z}$, then $I(X \to Y \mid \hat{\mathbf{z}}) = 1$.

In our application, we use the causal information flow to decide which variable should be added to the tree being built. This is shown in Algorithm 2. At line 2, we use the *do*-conditioning on the already build path $\mathbf{p}$, and we allow inputting additional observed variables $\mathbf{O}$ in an additional conditioning set. We replace $Y$ from (3) with the state of the explanandum $e$, suppress the corresponding summation and divide by the prior probability $p(e \mid \mathbf{o}, \hat{\mathbf{p}})$, so that we end up computing $I(X \to e \mid \mathbf{o}, \hat{\mathbf{p}}) =$

$$\sum_{x \in \mathcal{X}} \frac{p(x \mid \mathbf{o}, \hat{\mathbf{p}}) p(e \mid \mathbf{o}, \hat{x}, \hat{\mathbf{p}})}{p(e \mid \mathbf{o}, \hat{\mathbf{p}})} \log \frac{p(e \mid \mathbf{o}, \hat{x}, \hat{\mathbf{p}})}{\sum_{x' \in \mathcal{X}} p(x' \mid \mathbf{o}, \hat{\mathbf{p}}) p(e \mid \mathbf{o}, \hat{x}', \hat{\mathbf{p}})},$$

ensuring that the expected value $\mathbb{E}_E\big[I(X \to e \mid \mathbf{o}, \hat{\mathbf{p}})\big] = \sum_{e \in \mathcal{E}} p(e \mid \mathbf{o}, \hat{\mathbf{p}}) I(X \to e \mid \mathbf{o}, \hat{\mathbf{p}})$ equals $I(X \to E \mid \mathbf{o}, \hat{\mathbf{p}})$.

Using this criterion, the explanation tree is then built as follows: the root node is selected as being $\arg\max_X I(X \to e \mid \mathbf{o})$; i.e., the node which has the maximum information flow to the state of the explanandum. The important part is that we may condition on $\mathbf{o}$ without confusing observation and explanandum. Furthermore, we also allow selection of an observed variable $X \in \mathbf{O}$ in addition to unobserved variables, consistently with our desiderata. When $X \in \mathbf{O}$, it is observed and we know its value $x$. We must then compute the pointwise causal information flow from $x$ to $e$, with $\mathbf{o}'$ being $\mathbf{o}$ without the observation $X = x$:

$$I(x \to e \mid \mathbf{o}', \hat{\mathbf{p}}) = \log \frac{p(e \mid \mathbf{o}', \hat{\mathbf{p}}, \hat{x})}{\sum_{x' \in \mathcal{X}} p(x' \mid \mathbf{o}', \hat{\mathbf{p}}) p(e \mid \mathbf{o}', \hat{x}', \hat{\mathbf{p}})}.$$

The tree is then grown recursively: for each possible value for $X$, a branch is added to the root. For each new leaf, the next explanatory variable is selected as being $\arg\max_Y I(Y \to e \mid \mathbf{o}, \hat{x})$, and so on, where the *do*-conditioning set always reflects the selected variable values from the root to the current leaf. We use only one stopping criterion, the minimum information flow $\alpha$ we accept as a causal information contribution. The algorithm furthermore allows explicitly to restrict the search set for explanatory variables $\mathbf{H}$ (defaulting to $\mathbf{V} \setminus \{E\}$). Finally, each leaf is labeled with $\log\big(p(e \mid \mathbf{o}, \hat{\mathbf{p}}) / p(e \mid \mathbf{o})\big)$ (where we make sure that variables selected in $\hat{\mathbf{p}}$ are removed from $\mathbf{o}$ if needed). This measures how much performing the interventions $\hat{\mathbf{p}}$ changes the probability of the explanandum (given the observations) with respect to the prior probability of the explanandum. Higher values indicate better explanations; negative values indicate that the probability of the explanandum actually decreases with the proposed explanation.

Using the information flow criterion brings us two advantages over standard (conditional) mutual information: first, we automatically only consider variables that can causally influence the explanandum. Second, when selecting the $i$th variable on a tree branch, we take into account the previously selected variables 1 through $i - 1$ causally, as they enter the conditioning set of variables that have been intervened on.

In practice, computing a causal information flow of the type $I(X \to e \mid \mathbf{o}, \hat{\mathbf{p}})$ at line 2 of Algorithm 2 requires

**Algorithm 2** Causal Explanation Tree

**function** $T = \text{CausalExplTree}(\mathbf{H}, \mathbf{o}, e, \hat{\mathbf{p}};\ \alpha)$

| | | |
|---|---|---|
| **Input:** | $\mathbf{H}$ : | set of explanatory variables |
| | $\mathbf{O} = \mathbf{o}$ : | observation set |
| | $E = e$ : | explanandum |
| | $\hat{\mathbf{p}}$ : | path of interventions |
| | $\alpha$ : | stopping criterion |
| **Output:** | $T$ : | a causal explanation tree |

$X^* \leftarrow \arg\max_{X \in \mathbf{H}} I(X \rightarrow e \,|\, \mathbf{o}, \hat{\mathbf{p}})$
**if** $I(X^* \rightarrow e \,|\, \mathbf{o}, \hat{\mathbf{p}}) < \alpha$ **then return** $\emptyset$

$T \leftarrow$ new tree with root $X^*$
**for each** $x \in \text{domain}(X^*)$ **do**
  $T' \leftarrow \text{CausalExplTree}(\mathbf{H} \setminus \{X^*\}, \mathbf{o}, e, \hat{\mathbf{p}} \cup \{\hat{x}\})$
  add a branch $x$ to $T$ with subtree $T'$ and
  assign it the contribution $\log\big(p(e \,|\, \mathbf{o}, \hat{\mathbf{p}}, \hat{x}) / p(e \,|\, \mathbf{o})\big)$
**end for**
**return** $T$

to know the distributions $p(X \,|\, \mathbf{o}, \hat{\mathbf{p}})$, $p(E \,|\, \mathbf{o}, \hat{\mathbf{p}})$, and $p(E \,|\, \mathbf{o}, \hat{X}, \hat{\mathbf{p}})$ (i.e., $p(E \,|\, \mathbf{o}, \hat{x}, \hat{\mathbf{p}})$ for all $x \in \mathcal{X}$). Additionally, $p(e \,|\, \mathbf{o})$ is needed to label the leaves, but as it is only dependent on $e$ and $\mathbf{o}$, we compute it only once. We can further avoid unnecessary computations by using the graphical reachability criterion from a candidate node $X$ to $E$, blocking paths going through $\mathbf{O} \cup \mathbf{P}$. The inference steps were implemented using the factor graph message-passing algorithm (Frey et al., 2001).

The complexity of this algorithm, in terms of number of calls to an inference engine per node in the constructed tree, is $\mathcal{O}(nd)$, where $n$ is the number of explanatory variables $|\mathbf{H}|$ and $d$ is the average domain size of the variables, e.g., 2 for binary variables. For comparison, Flores's (2005) approach is $\mathcal{O}(n^2 d^2)$.

## 5 EXPERIMENTS

We compare causal explanation trees (CET) with parameter $\alpha = 0$ to Most Probable Explanation (MPE), Bayes' factor (BF) following Yuan and Lu (2007), and standard (noncausal) explanation trees (ET) with parameters $\alpha = 0.02$ and $\beta = 0$. We test the approach on three simple networks[3] to compare the relevance of explanations. A more extended version of these experiments and comments can be found in Nielsen (2007).

**Drug** (Figure 1). This network comes from Pearl (2000, chap. 6). It represents the outcome of an experiment designed to check the efficiency of a new drug on male and female patients. The males have a natural recovery rate of 70%; taking the drug decreases it to 60%. Similarly, 30% of females recover naturally, but only 20% when given the drug. Thus, both the absence of drug and being a male can explain a good recovery rate.

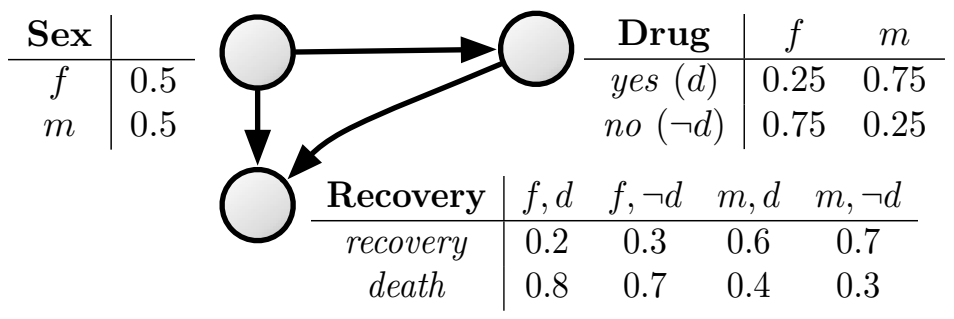

Figure 1: The Drug network.

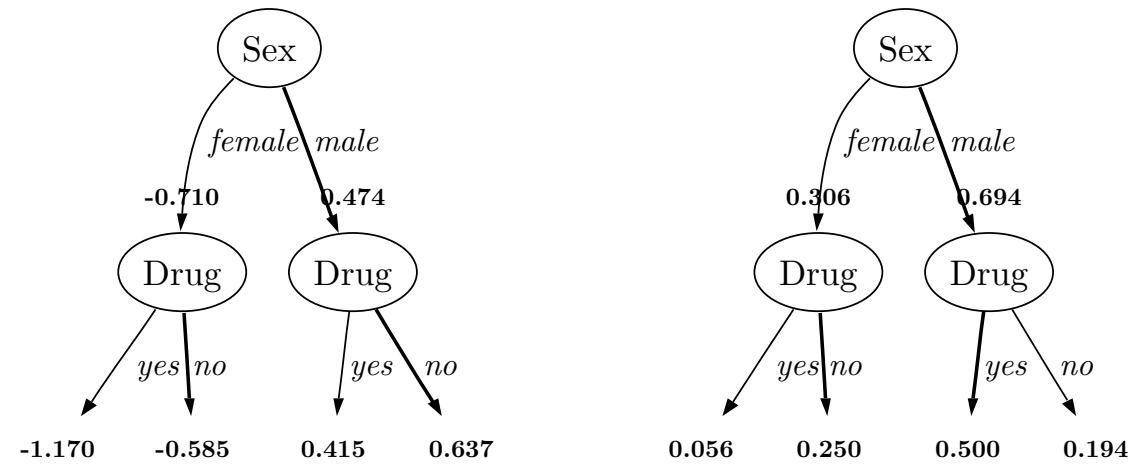

(*a*) Causal explanation tree (*b*) Explanation tree

(*c*) MPE: $p(\mathit{Sex} = m,\ \mathit{Drug} = \mathit{yes} \,|\, \mathit{Recovery} = \mathit{rec.}) = 0.5$
(*d*) BF: $\text{BF}(\mathit{Sex} = m) = 2.27$
$\text{BF}(\mathit{Sex} = m,\ \mathit{Drug} = \mathit{no}) = 1.68$
$\text{BF}(\mathit{Drug} = \mathit{yes}) = 1.25$

Figure 2: Drug: explain *Recovery* = *recovery*.

In Figure 2, we try to explain a recovery. All approaches correctly realize that *Sex* = *m* largely accounts for the recovery. However, ET selects *Sex* = *m* ∧ *Drug* = *yes* as the best explanation according to the leaves' labels, just like MPE. This contradicts the natural idea of explanation, since the drug has a negative impact on the recovery. CET labels the leaves more sensibly: branches where the drug was not given have a higher rank. Moreover, the branches where *Sex* = *f* have a negative label, indicating that they actually decrease the probability of recovery. Although the first two BF explanations are sensible, the third one is mistakenly selects *Drug* = *yes* as an explanation.

**Academe** (Figure 3). This network depicts the relationships between various marks given to students following a course. The *Final mark* is determined by some *Other* outside factors and an intermediate mark (*T.P. mark*), which is in turn determined by the student's abilities in *Theory* and *Practice* as well as *Extra* curricular activities in this tested subject.

In Figure 4, the explanandum was set to *Final mark* = *fail*; i.e., we want to explain why a student failed the course. *T.P. mark* and *Global mark* have been excluded from the possible explanatory variables in the two tree algorithms as they are modeling artifacts.

ET tells us that *Theory* = *bad* is the best explanation. We could have expected *Practice* to also be part of

[3]The conditional probability tables have been omitted in the figures of the two larger BNs due to lack of space and can be found at `http://www.zurich.ibm.com/~uln/causalexpl/`.

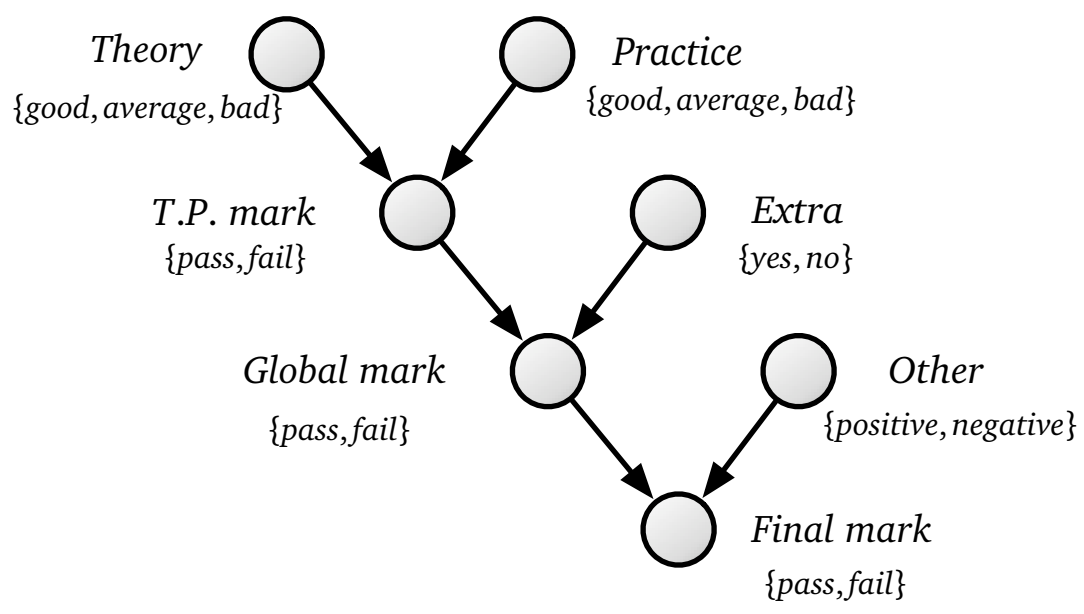

Figure 3: The ACADEME network.

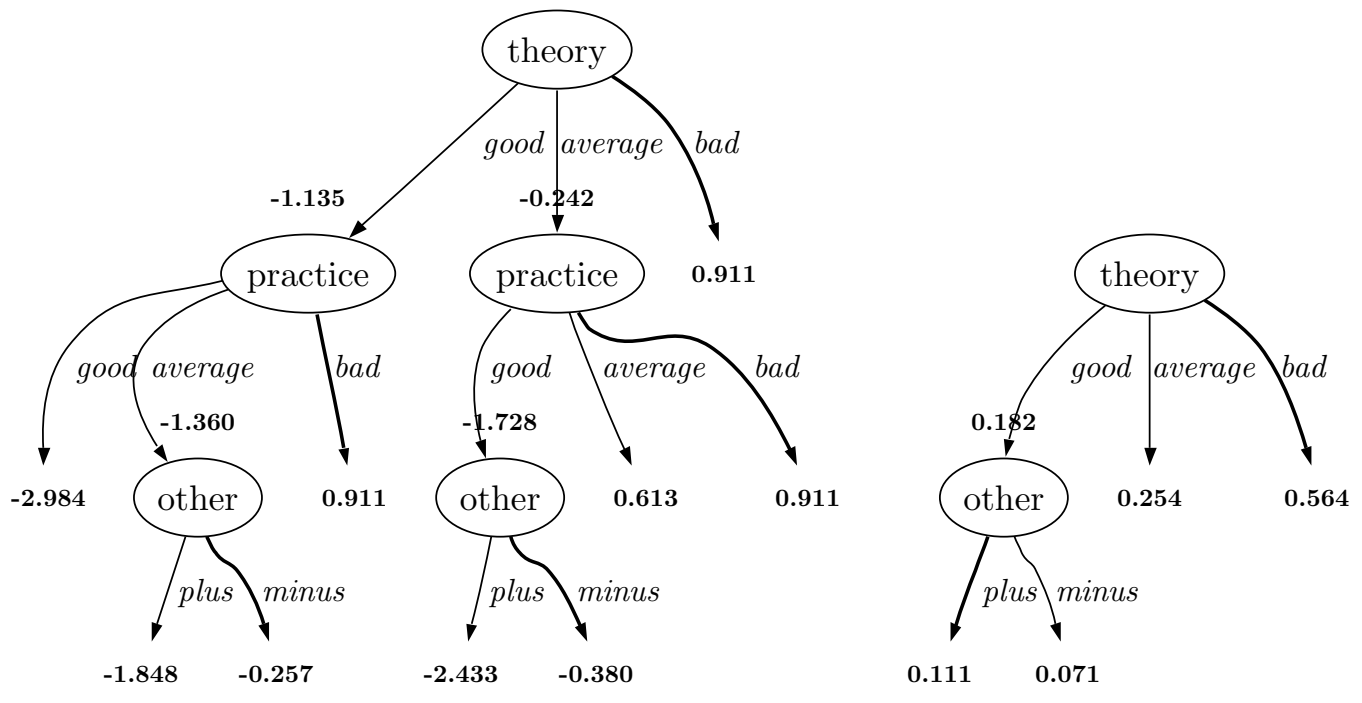

(*a*) Causal explanation tree (*b*) Explanation tree

(*c*) MPE: $p(\textit{Theory} = \textit{bad},\ \textit{T.P. mark} = \textit{fail},\ \textit{Global mark} = \textit{fail},\ \textit{Extra} = \textit{no},\ \textit{Other} = \textit{positive},\ \textit{Practice} = \textit{good} \mid \textit{Final mark} = \textit{fail}) = 0.208$

(*d*) BF: $\text{BF}(\textit{Theory} = \textit{bad}) = 3.02$
$\text{BF}(\textit{Theory} = \textit{bad},\ \textit{Extra} = \textit{no}) = 2.78$
$\text{BF}(\textit{Theory} = \textit{bad},\ \textit{Other} = \textit{negative}) = 2.53$

Figure 4: ACADEME: explain *Final mark* = *fail*.

alternate explanations, as it influences the final mark very similarly to *Theory*. This is what CET does, including *Practice* to explain the final failure when *Theory* is *average* or *good*. MPE includes *Practice* = *good* in its long list of states, which does not seem intuitively likely. BF provides more concise explanations, but, like ET, ignores *Practice* altogether, although a bad practice can account for failure equally well.

**Asia** (Figure 5). This network (Lauritzen and Spiegelhalter, 1988) models the relationships between two indicators, X-ray results and dyspnea, of severe diseases for a person. Tuberculosis (more likely if a visit to Asia occurred) and lung cancer (more likely when the person smokes) both increase abnormal X-ray results and dyspnea; bronchitis also causes increased dyspnea. *TbOrCa* is a modeling artifact, excluded from the the analysis in the two tree algorithms.

In Figure 6, we try to explain abnormal X-ray results. Whereas both tree algorithms select a *Lung cancer* as the best explanation, they differ on how to explain when it is *absent*: CET selects, justifiably, *Tuberculosis*, but ET uses *Dyspnea* and then *Bronchitis*, which are not causes of *X-ray* and cannot explain it, especially not when we know that no lung cancer is present. MPE surprisingly excludes a visit to Asia, when this is expected to make abnormal X-rays more likely through *Tuberculosis*. BF, while still providing very concise explanations, is stuck on the middle node *TbOrCa* and, like ET, ignore the important *Tuberculosis*.

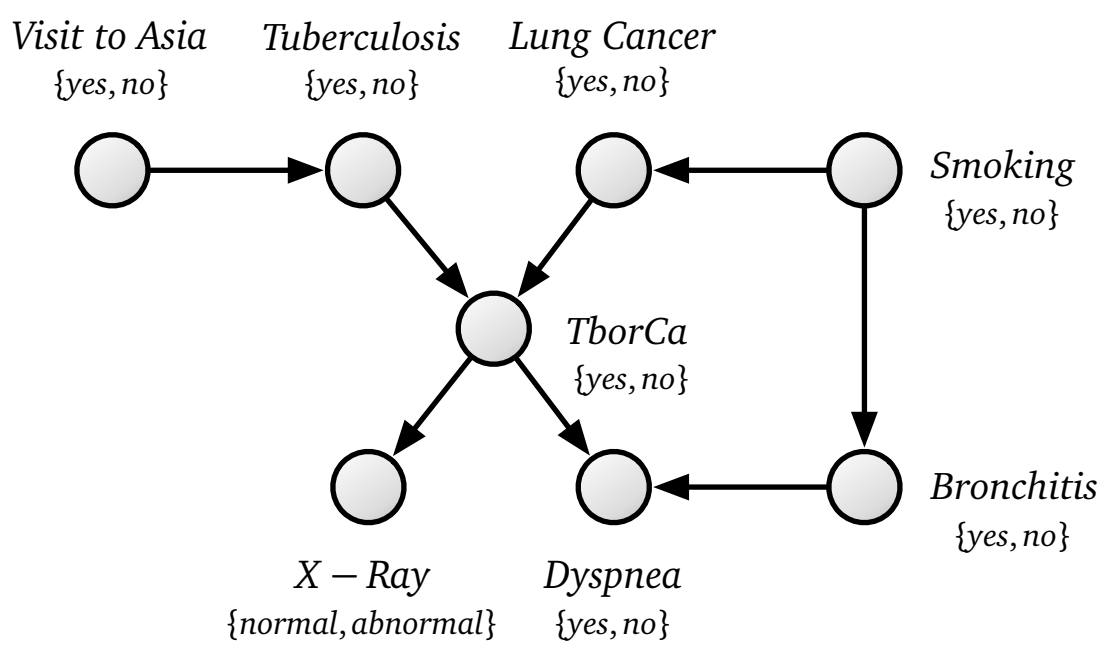

Figure 5: The ASIA network.

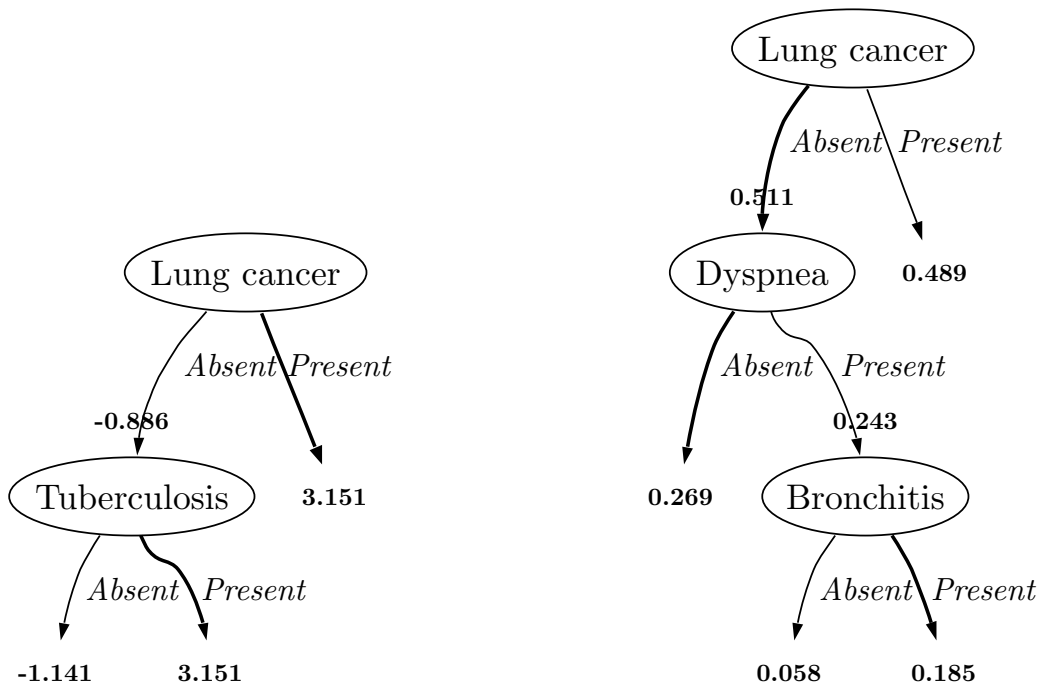

(*a*) Causal explanation tree (*b*) Explanation tree

(*c*) MPE: $p(\textit{Bronchitis} = \textit{yes},\ \textit{Dyspnea} = \textit{yes},\ \textit{Lung cancer} = \textit{yes},\ \textit{TbOrCa} = \textit{yes},\ \textit{Smoker} = \textit{yes},\ \textit{Tuberculosis} = \textit{yes},\ \textit{Visit to Asia} = \textit{no} \mid \textit{X-ray} = \textit{abnormal}) = 0.24$

(*d*) BF: $\text{BF}(\textit{TbOrCa} = \textit{yes}) = 19.60$
$\text{BF}(\textit{TbOrCa} = \textit{yes},\ \textit{Visit to Asia} = \textit{no}) = 19.21$
$\text{BF}(\textit{TbOrCa} = \textit{yes},\ \textit{Lung cancer} = \textit{yes}) = 16.42$

Figure 6: ASIA: explain *X-ray* = *abnormal*.

In Figure 7, we try to explain the presence of dyspnea for a smoker. While CET is still able to select *Smoker* as explanatory variable, ET can only add this observation to the explanandum and thus cannot select it. Instead, the best explanation according to ET is normal X-rays, which does not seem very likely. Although ET does select the important *Lung cancer* and *Tuberculosis*, it ignores the largest factor according to BF and CET, namely *Bronchitis*. Here too, CET selects the more intuitively interpretable explanations.

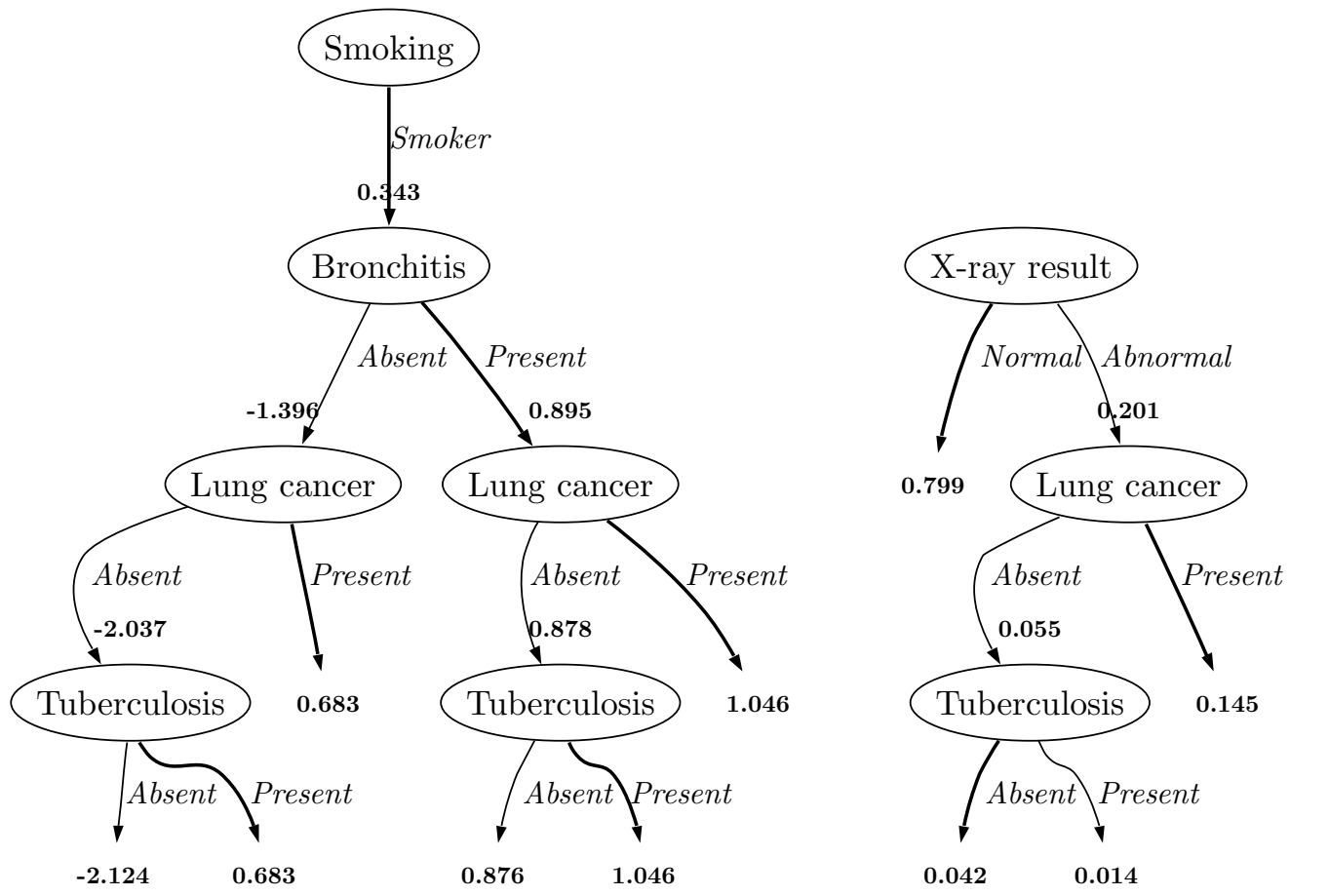

($a$) Causal explanation tree ($b$) Explanation tree

($c$) MPE: $p$(*Bronchitis* = *yes, X-ray* = *normal, Lung cancer* = *no, TbOrCa* = *no, Smoker* = *yes, Tuberculosis* = *no, Visit to Asia* = *no* | *Dyspnea* = *yes*) = 0.46

($d$) BF: BF(*Bronchitis* = *yes*) = 6.14
BF(*Bronchitis* = *yes, Visit to Asia* = *no*) = 5.89
BF(*Bronchitis* = *yes, Tuberculosis* = *no*) = 5.84

Figure 7: ASIA: explain *Dyspnea* = *yes* | *Smoker* = *yes*.

## 6 CONCLUSION

We have presented an approach to explanation in causal BNs, causal explanation trees. Explanations are presented as a tree, compactly representing several explanations and making it more readable than a (possibly long) list. Assuming that the BN is causal allows us to use the causal information flow criterion to build the tree. This leads to more sensible explanations, in that we only explain a given state with variables that can causally influence it. The approach makes an explicit distinction between observation and explanandum. This lets the user input all available knowledge about the network as observation, while still focusing on explaining one of them and allowing the observed variables to be selected as part of a good explanation. The algorithm labels the leaves so as to reflect how a proposed explanation changes the probability of the explanandum, making the tree easy to interpret.

Causal explanation trees, unlike other techniques, do not condition on the explanandum to maximize the probability of the explanatory variables $p(\mathbf{h} \,|\, \mathbf{e})$, but focus on $p(\mathbf{e} \,|\, \mathbf{h})$ instead by means of the causal information flow. This allows it to compare favorably to MPE, Bayes' factor, and Flores's (2005) noncausal explanation trees on the tested networks because the returned explanations are intuitive, and those with a positive label in the tree ensure that they increase the probability of the explanandum.